\documentclass[]{elsarticle}
\graphicspath{ {./figures/} }
\usepackage{hyperref}
\usepackage{float}
\usepackage{verbatim} %comments
\usepackage{apalike}
\restylefloat{figure}
\floatstyle{plaintop} %table caption at top
\restylefloat{table}

\usepackage{booktabs}
\usepackage{amssymb}
\usepackage{pifont}

\newcommand{\xmark}{\ding{55}} % cross
\newcommand{\cmark}{\checkmark} % tick

\makeatletter
\def\ps@pprintTitle{%
 \let\@oddhead\@empty
 \let\@evenhead\@empty
 \let\@oddfoot\@empty
 \let\@evenfoot\@empty
}

\usepackage{tikz}
\usetikzlibrary{positioning, shapes.geometric, arrows.meta, fit}

\tikzstyle{mainagent} = [circle, minimum size=2.5cm, draw=black, thick, align=center]
\tikzstyle{subagent} = [circle, minimum size=2cm, draw=black, thick, align=center]
\tikzstyle{arrow} = [->, thick, >=stealth]
\biboptions{authoryear}

\begin{document}
	\begin{frontmatter}
		
		%The title page must contain the title of the paper and the full name/full affiliation with country/e-mail address for each author and co-author of the manuscript. Please make sure you have included all elements listed below with your manuscript submission.
		
        \begin{titlepage}
        	\begin{center}
        		\vspace*{1cm}
        		
        		\textbf{ \large Chatting with your ERP: A Recipe}
        		
        		\vspace{1.5cm}
        		
        		% Author names and affiliations
        		Jorge Ruiz Gómez$^a$\textsuperscript{*}(jorge.ruiz@itcl.es)\\,
                    Lidia Andrés Susinos$^a$ (lidia.andres@itcl.es)\\,
                    Jorge Álamo Olivé$^a$ (jorge.alamo@itcl.es)\\
                    Sonia Rey Osorno$^a$ (sonia.rey@itcl.es)\\
        		Manuel Luis Gonzalez Hernández$^a$ (manuel.gonzalez@itcl.es)\\
        		
        		\hspace{10pt}
        		
        		\begin{flushleft}
        			\small  
        			$^a$ Instituto Tecnológico de Castilla y León, Burgos, Spain \\

        			\vspace{1cm}
        			\textbf{Corresponding author at: Instituto Tecnológico de Castilla y León, Burgos, Spain} \\
        			Jorge Ruiz Gómez \\
        			Instituto Tecnológico de Castilla y León, Burgos, Spain \\
        			Tel: +34 947 29 84 71 \\
        			Email: jorge.ruiz@itcl.es
        			
        		\end{flushleft}        
        	\end{center}
        \end{titlepage}
		
		\title{Chatting with your ERP: A Recipe}
		
		\author[label1]{Jorge Ruiz Gómez \corref{cor1}}
		\ead{jorge.ruiz@itcl.es}
		
		\author[label1]{Lidia Andrés Susinos}
		\ead{lidia.andres@itcl.es}
		
		\author[label1]{Jorge Álamo Olivé }
		\ead{jorge.alamo@itcl.es}
        
		\author[label1]{Sonia Rey Osorno}
		\ead{sonia.rey@itcl.es}

		\author[label1]{Manuel Luis Gonzalez Hernández }
		\ead{jorge.alamo@itcl.es}
		\cortext[cor1]{Corresponding author.}
		\address[label1]{Instituto Tecnológico de Castilla y León, Burgos, Spain}
		
		\begin{abstract}
        This paper presents the design, implementation, and evaluation behind a Large Language Model (LLM) agent that chats with an industrial production-grade ERP system. The agent is capable of interpreting natural language queries and translating them into executable SQL statements, leveraging open-weight LLMs. A novel dual-agent architecture combining reasoning and critique stages was proposed to improve query generation reliability.
        
		\end{abstract}
		
		
	\end{frontmatter}
	
	\section{Introduction}
	\label{introduction}

        Enterprise Resource Planning (ERP) systems are complex software platforms that integrate and manage core business processes across departments such as manufacturing, logistics, finance, and human resources. These systems are essential for coordinating operations, ensuring data consistency, and enabling data-driven decision-making in industrial environments. However, the complexity and technical nature of ERP databases often require users to have specialized knowledge of database schemas to retrieve relevant information. 

        Recent advancements in artificial intelligence (AI), specially in the field of Large Language Models (LLMs), have created new possibilities for enhancing the interaction between users and  industrial manufacturing workflows. This interaction can be developed thanks to LLMs being capable of interpreting human-like language and generating human-like responses, and this generation can be accomplished while having access to complex data sources, bridging the gap between natural language and structured data systems.
        
        This work introduces an intelligent software agent designed to interface with the ERP of a defense manufacturing company. This agent leverages open-weight LLMs to translate user queries expressed in natural language into executable SQL statements, enabling intuitive and efficient access to ERP data. To enhance robustness and reliability, the system incorporates a dual-agent architecture combining reasoning and critique components. This structure allows for iterative refinement and validation of generated queries, significantly improving their accuracy and relevance based on a partial result analysis.

	\section{Related Work}
	\label{title_page}
    
        The task of converting Natural Language into SQL, commonly known as Text-to-SQL (Text2SQL), has witnessed rapid development in the latest years thanks to the widespread adoption of LLMs.  

        Early approaches to solve this problem relied on rule-based systems \cite{10.14778/2735461.2735468} and supervised learning with DNN and LSTM architectures \cite{lstm-10.1162/neco.1997.9.8.1735}.

        Recent attention has shifted towards leveraging LLMs as zero- or few-shot agents capable of performing semantic parsing through prompt-based interaction, thus circumventing the need for task-specific models.

        This paradigm shift has catalyzed a new line of research focused on LLM agents, which integrate LLMs with external tools and structured reasoning workflows to improve SQL generation performance without modifying model parameters.

        Initial iterations of LLM-based Text-to-SQL demonstrated the surprising effectiveness of zero- and few-shot LLMs such as GPT-3 and GPT-4 in generating syntactically valid and semantically accurate SQL from natural language \cite{rajkumar2022evaluatingtexttosqlcapabilitieslarge}.
        \cite{liu2023comprehensive} \cite{gao2023texttosql}.

        These single-pass prompting strategies exhibited limitations in handling complex queries, ambiguous schemas, or large-scale relational contexts, especially in cross-domain or enterprise-level settings.

        To address these challenges, recent studies \cite{xie2024magsqlmultiagentgenerativeapproach} \cite{cen2025sqlfixagentsemanticaccuratetexttosqlparsing} have introduced LLM-based agentic architectures that include multi-step reasoning, execution feedback, schema understanding and query refinement, imitating human-like problem solving workflows without the need of fine-tuning the language models. 

        Additionally, the use of chain-of-thought reasoning and least-to-most prompting also integrates well into the agent paradigm, allowing for hierarchical decomposition of complex user intent into compositional SQL fragments.

        This agentic designs that are based on pretrained LLMs seem to offer several advantages. First,  the agentic design mitigate the dependency on large annotated datasets, secondly, enhance the results through modular reasoning, and last, increase the adaptability across unfamiliar schemas and unknown domains without the need of retraining the model. However, agentic architectures often face significant challenges, such as high inference latency due to iterative LLM calls and brittle orchestration pipelines that rely heavily on very specific prompts, designed to work well with a handful of models and loss of information, due to the agent not sharing the reasoning behind their decisions and actions.

        \section{Architecture}
        \label{Arcitecture}

        Our chatbot driven Text2SQL architecture implements a REACT-Based (Reason + Act)  conversational agent that can query a database with natural language via a second agent, the SQL Agent.  The REACT framework enables the agent to combine iterative reasoning with action execution. The architecture follows a modular design, where reasoning, execution, and evaluation are handled by distinct components. A Human-in-the-Loop (HITL) mechanism is included to ensure clarity and correctness before executing any database queries.

        \subsection{REACT Agent}

        The system's entry point is the REACT agent. Upon receiving user input, the agent initiates a multi-step process: it first interprets the user’s intent, then determines whether the intent can be addressed through a database query. If so, the REACT agent delegates the task to the SQL agent.
        
        The SQL agent processes the query and returns the query results to the REACT agent, which then performs reasoning over the output and answers the user with a natural language response.

        \subsection{SQL Agent}

        The SQL Agent transforms natural language queries into optimized SQL queries. The core design of this agent is a two-agent collaborative loop that is designed to generate an optimal and refined query.
        The first agent in this loop, the \textbf{the reasoner}, generates, executes and improves SQL Queries based on the user request and feedback. The second agent, the \textbf{critic}, validates the query, the results and provides feedback to refine it. 
        This cooperative process ensures syntactic correctness, semantic adequacy, and usability of the generated queries, even in complex database scenarios.

        Both agents share a message state and behave as if they are participating in a collaborative conversation, exchanging structured messages to iteratively improve the SQL query.

        \subsubsection{SQL Reasoner}
        The Reasoner agent is responsible for the initial comprehension of user queries and the generation of candidate SQL statements. It implements a self-debugging strategy inspired by \cite{chen2023teachinglargelanguagemodels}).
            
        To begin, the agent constructs a preliminary SQL query based on the user's intent and an internal Markdown-based representation of the database schema. It then attempts to execute this initial query.
        
        If execution fails—due to syntax errors, undefined identifiers, or improper joins—the agent analyzes the resulting error message and iteratively refines the query to resolve the issue.
        
        This stage aims to address common low-level errors, such as malformed syntax, invalid field references, and typographical mismatches in table or column names. The Reasoner can generate up to $N$ attempts per query to produce an executable SQL statement. 
        
        At this stage, the agent lacks the ability to assess the semantic adequacy or pragmatic usefulness of the result set. Consequently, it is complemented by a dedicated \textbf{Critic agent}.

        \subsubsection{SQL Critic} 
        
        The Critic agent functions as an internal evaluator of the SQL queries produced by the Reasoner. Its role is to assess the quality, correctness, and efficiency of the query output according to several formal and practical criteria. 

        First, it ensures syntactic validity by verifying that the SQL query is a well-formed SELECT statement. 
        Second, it evaluates semantic appropriateness, determining whether the query responds to the user’s intent by considering table relationships, column semantics, and logical constraints. In this step, the agent also reviews the readability of the query, making sure that the columns are meaningful and related to the user intent. 
        Third, the critic reviews query efficiency, identifying redundant operations, unnecessary joins, or overly complex subqueries that could hinder performance. 

        Lastly, the agent reviews a subset of the results, checking that they conform the user intent. 

        When deficiencies are detected, it returns structured feedback to the Reasoner, initiating a refinement cycle. This iterative dialogue between Critic and Reasoner is limited to a maximum of $M$ rounds. 

        
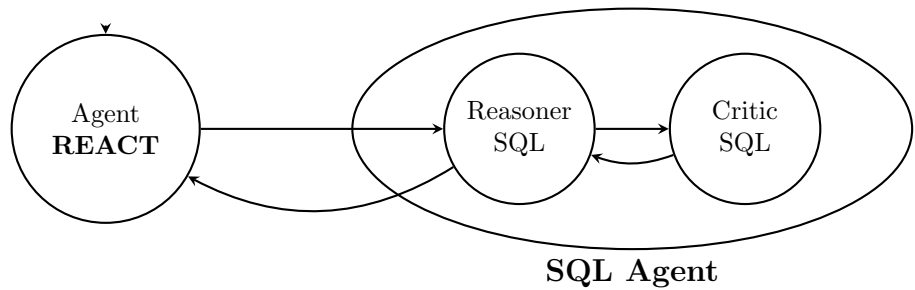
\vspace{1.5cm}
        \begin{tikzpicture}[node distance=2cm and 2cm, on grid]
        \label{graph-diagram}
        % Nodes
        \node[mainagent] (react) {Agent\\\textbf{REACT}};
        \node[subagent, right=5.5cm of react] (reasoner) {Reasoner\\SQL};
        \node[subagent, right=3cm of reasoner] (critic) {Critic\\SQL};
        
        % Group: SQL Agent (ellipse around both)
        \node[ellipse, draw=black, thick, fit=(reasoner)(critic), label=below:{\large \textbf{SQL Agent}}] (sqlcontainer) {};
        
        % Arrows
        \draw[arrow] (react) -- (reasoner);
        \draw[arrow] (reasoner) -- (critic);
        \draw[arrow] (critic) to[bend left=20] (reasoner);
        \draw[arrow] (reasoner) to[bend left=30] (react);

        % Optional: user emoji above
        \node[above=1.5cm of react] (user) {};
        \draw[arrow] (user) -- (react);
        
        \end{tikzpicture}

        \subsection{Database Schema}

        To ensure context-aware SQL query generation, a structured schema of the database is injected into the SQL Agent. This schema acts as a bridge between the agent and the underlying data structures, improving the agent contextual understanding of the problem. 

        The schema, encoded as markdown, is organized into two well differentiated sections. The first one, is a handcrafted semantic description of the tables and its relations by an expert developer, while the second one is and auto-generated list of tables and columns. 

        \subsubsection{Semantic Description}
        \label{schema-handcrafted}
        The first section of the schema is a carefully curated natural language description of the tables, their semantic representation, their most important columns and their relations. 

        This section helps the agent to connect the concepts expressed by the users in natural language with the internal structure of the database, as no every concept has a direct representation in the schema, and automatic schema descriptions are often ambiguous or technical. 

        This section is organized into the following subsections:

        \begin{itemize}
            \item \textbf{Introduction:} A general description of the role and purpose of the database
            \item \textbf{Concepts:} Explanations of core domain entities and how they are related.
            \item \textbf{Table Summaries:} A concise description of each relevant table, highlighting its purpose, key columns, and relationships with other tables.
            \item \textbf{High-Level Relationships}: A set of manually defined foreign-key-like mappings that are not formally declared within the database. 
        \end{itemize}

        \subsubsection{Autogenerated Schema}
        Complementing the expert-authored layer is an automatically extracted schema that includes a comprehensive enumeration of all tables, columns, data types, key constraints, and known relationships.

        For each row, a markdown list item is created with the following fields:
        
        \begin{itemize}
          \item Column name and data types
          \item Column description, if available
          \item Sample values to illustrate typical field contents
          \item Primary and foreign key indicators
          \item Explicit table-to-table relationships, if available
        \end{itemize}

        \subsection{Human in the Loop}
        To ensure safety, interpretability and alignment with the user's intent, the REACT agent implements a human in the loop interaction mechanism that tries to clarify the user intent.

        An LLM analyzes the user intent and tries to prevent incorrect or overly vague natural language prompts that might waste precious compute time building incorrect queries. 

        If the intent is clear, specific, can be interpreted without confusion and answerable with the information present in the schema, then the intent does not require additional information from the user and a SQL Query is iterative generated, otherwise the user is asked to clarify its intent. 

        \subsection{Reasoned Structured Outputs}
        To ensure compatibility with LLMs that do not support structured outputs or tool calling, while maintaining those functionalities, the system implements a hybrid reasoning-extraction pipeline that decouples the LLM reasoning from the less-creative structured output generation. 

        In the first step, an LLM is prompted to reason and solve a task. This model is not required to support structured outputs, however, it must present the partial or final results, generated through the reasoning process, within markdown code blocks.  This format encourages human-like interaction, a feature which most instruct models excel at. 

        In a second step, an smaller LLM with structured output support is asked to extract the most relevant reasoned markdown block, matching an specific structured schema.  

        \section{Experimental Set-Up}
        This section outlines the methodology used to execute the SQL Agent designed to interact with a production-ready relational database.

        \subsection{Database}
        The experiments were performed within a production-grade SQL Server 2017 database connected with an ERP. This SQL server contains 7 tables and 321 columns, as shown in \ref{tab-db-columns}. 
        
        As the database contains sensitive data, by client request, all the data in this document will be anonymized. 

        All of these tables are related, but only one foreign key is defined as shown in \ref{tab-relations}. Only 119 columns are NLP documented, and none of the comments contain references to other fields. 

        As stated in \ref{schema-handcrafted}, an expert developer has crafted a semantic description of the database schema to enhance the agent’s understanding of the data structure. 

        \begin{table}[H]
        
        \centering
        \begin{tabular}{|l|c|}
        \hline
        \textbf{Table Name} & \textbf{Number of Columns} \\
        \hline
        T\_A & 28 \\
        T\_B & 77 \\
        T\_C & 15 \\
        T\_D & 37 \\
        T\_E & 43 \\
        T\_F & 67 \\
        T\_G & 54 \\
        \hline
        \end{tabular}
        \caption{\label{tab-db-columns}Anonymized table structure with placeholder column counts}
        \end{table}

        \begin{table}[H]
        
        \centering
        \begin{tabular}{|l|l|l|l|}
        \hline
        \textbf{Source Table} & \textbf{Related Table} & \textbf{Relationship Type} & \textbf{Linking Key / Comment} \\
        \hline
        \textbf{T\_A}          & T\_B                  & 1 to N                    & \texttt{idA}                                 \\
        \textbf{T\_A}          & T\_C                  & 1 to 1                    & \texttt{idA}                                  \\
        \textbf{T\_A}          & T\_D                  & 1 to N (implicit)         & via \texttt{CurrentCode} → \texttt{UnitNumber} \\
        \textbf{T\_A}          & T\_G                  & 1 to N (implicit)         & via \texttt{CurrentCode} → \texttt{UnitNumber} \\
        \textbf{T\_B}          & T\_E                  & 1 to 1                    & \texttt{idB}                               \\
        \textbf{T\_C}          & T\_E                  & 1 to N                    & \texttt{idC}                                \\
        \textbf{T\_D}          & T\_F                  & 1 to N                    & \texttt{ID} → \texttt{PathID} (Foreign Key)  \\
        \textbf{T\_G}          & T\_F                  & 1 to 1                    & \texttt{UnitNumber} → \texttt{UnitNumber}     \\

        \hline
        \end{tabular}
        \caption{\label{tab-relations}Anonymized relationship mapping between database tables}
        \end{table}

        \subsection{Expert Validation}
        The agent was evaluated using a set of twelve distinct queries, each curated by an experienced ERP system user to reflect realistic and domain-relevant information needs. Given the variability in query outcomes, due to the use of aliases or dynamic column selection, each response was validated by an expert.

        \subsection{LLM Models and Prompts}
        Due to the data privacy policy, the experiments were performed using eleven different open-weight LLMs.  Each model was employed for the tasks of sql-reasoning and sql-critic. Additionaly, Qwen2.5-32B was employed to handle multi-turn dialogue, while LLaMA 3.1-8B was used to extract structured outputs from the reasoning process. 

        All models were deployed using Ollama, with each instance configured to support a context window of 65,536 tokens and a maximum token generation time of three minutes. The evaluation was performed on a high-performance workstation equipped with an AMD Ryzen Threadripper 7960X processor featuring 24 cores, 126 GB of system memory, and three NVIDIA RTX 4090 graphics cards, each with 24 GB of dedicated VRAM.
        
        The prompts have been built around Qwen 2.5 32B, so it is expected that not every model will perform equally well. Differences in model architecture, training data, and instruction-following capabilities can lead to significant variations in how effectively each model interprets and responds to the same prompt, so a disadvantage is expected. 
        
        \section{Results}

        This section presents a comparative of the performance of the agent across a set of eleven models and eleven questions. \ref{tab-results} summarizes the correctness of each model's responses, where a checkmark (\cmark{}) denotes a correct answer and a cross (\xmark{}) indicates an incorrect one. The accuracy column reflects the total number of correct responses out of eleven.

        The evaluation encompasses models of varying scales and fields of expertise. Notably,  Devstral 24B Q4 model achieved the highest accuracy (10/11), followed closely by Qwen 2.5 32B Q4 (9/11), indicating strong performance in this benchmark setting. In contrast, several models—including Codestral 22B Q4 and Magistral 24B Q4—did not produce any correct answers, highlighting significant variability in performance across different architectures.
        
        \begin{table}[h!]
        \centering
        \begin{tabular}{lcccccccccccc}
        \toprule
        \textbf{Model} & \textbf{1} & \textbf{2} & \textbf{3} & \textbf{4} & \textbf{5} & \textbf{6} & \textbf{7} & \textbf{8} & \textbf{9} & \textbf{10} & \textbf{11} & Accuracy \\
        \midrule
        Llama 3.1 8B FP16         & \xmark & \cmark & \xmark & \xmark & \xmark & \xmark & \xmark & \xmark & \xmark & \xmark & \xmark & 1/11 \\
        Qwen 2.5 7B FP16          & \cmark & \xmark & \xmark & \xmark & \cmark & \xmark & \cmark & \xmark & \xmark & \xmark & \xmark & 3/11 \\
        Qwen 2.5 32B Q4           & \cmark & \cmark & \xmark & \cmark & \cmark & \cmark & \cmark & \cmark & \cmark & \xmark & \cmark & 9/11 \\
        \textbf{Devstral 24B Q4}  & \cmark & \cmark & \cmark & \cmark & \cmark & \cmark & \cmark & \cmark & \cmark & \xmark & \cmark & \textbf{10/11}\\
        Codestral 22B Q4          & \xmark & \xmark & \xmark & \xmark & \xmark & \xmark & \xmark & \xmark & \xmark & \xmark & \xmark & 0/11 \\
        Magistral 24B Q4          & \xmark & \xmark & \xmark & \xmark & \xmark & \xmark & \xmark & \xmark & \xmark & \xmark & \xmark & 0/11 \\
        Deepseek Coder 33B Q4     & \xmark & \xmark & \xmark & \xmark & \xmark & \xmark & \xmark & \xmark & \xmark & \xmark & \xmark & 0/11 \\
        Gemma 3 27B Q4            & \cmark & \xmark & \xmark & \xmark & \xmark & \xmark & \cmark & \xmark & \xmark & \xmark & \xmark & 2/11 \\
        Qwen 3 32B Q4             & \cmark & \cmark & \xmark & \xmark & \xmark & \xmark & \xmark & \xmark & \xmark & \xmark & \xmark & 2/11\\
        \bottomrule
        \end{tabular}
        \caption{Model performance on questions 1–11. \cmark{} = correct, \xmark{} = incorrect}
        \label{tab-results}
        
        \end{table}

        It should be noted that Qwen 3 32B Q4 and Llama 3.3 70B Q4 were also included in the evaluation but were unable to complete inference due to generation time constraints and hardware limitations. Specifically, their memory requirements exceeded 100 GB of VRAM, which led to execution failure due to the three minute request time limit.

        \section{Recipe for a SQL Agent in your ERP}
        To implement this SQL in an ERP, a few steps must be followed.

        The first step involves exposing the internal SQL database of the ERP in read-only mode. This allows the SQL agent to safely query the system without risking data corruption. The agent must have access to the schema an the underlying data. 

        Once access is established, the most relevant fields—those frequently used or carrying business-critical information—should be clearly documented. This documentation should bridge the semantic gap between the internal field names and the domain-specific concepts they represent. 

        After documenting the fields, the ERP expert creates a natural language description of the ERP. This description includes the domain-specific concepts that the ERP represents, table relationships, roles, and such. 

        The last step is LLM selection and configuration. In our design, every single agent mentioned can be configured separately. A two model approach seems to offer low latency with good performance. This approach includes a lightweight model that has the responsibility of extracting structured data from reasoned responses, and a larger one for the rest of the tasks.

        \section{Conclusion and Future Work}
        This work has explored the feasibility and practical challenges of integrating LLMs into industrial environments for natural language interaction with structured data systems such as ERPs.

        Through the development and iterative refinement of an LLM-based software agent, we have shown that open-weight models can be employed effectively in scenarios involving complex SQL query generation.
        
        However, two significant challenges constrain the agent capabilities.

        First, the agent needs manual alignment with the database through a manual natural language description written by an expert. This setup step limits the scalability and usability of the approach, preventing a plug-and-play agent..

        Second, the agent has difficulties understanding which columns has to select based on vague prompts. Identifying the columns based solely on the ID format or without detailing what an id refers to relies on the model choosing the correct columns. 

        Also, the REACT agent sometimes fails to summarize the entire user intent to the SQL Agent, so some details or corrections might be skipped. 

        Future work will focus on solving both issues. First, a new agent to describes the database is proposed, this agent should reduce or even stop the need of a database expert to write down a natural language description of the database.  Second, we will investigate improved methods for intent interpretation and column selection under uncertainty.

	\bibliography{sample}

\begin{thebibliography}{8}
\expandafter\ifx\csname natexlab\endcsname\relax\def\natexlab#1{#1}\fi
\providecommand{\url}[1]{\texttt{#1}}
\providecommand{\href}[2]{#2}
\providecommand{\path}[1]{#1}
\providecommand{\DOIprefix}{doi:}
\providecommand{\ArXivprefix}{arXiv:}
\providecommand{\URLprefix}{URL: }
\providecommand{\Pubmedprefix}{pmid:}
\providecommand{\doi}[1]{\href{http://dx.doi.org/#1}{\path{#1}}}
\providecommand{\Pubmed}[1]{\href{pmid:#1}{\path{#1}}}
\providecommand{\bibinfo}[2]{#2}
\ifx\xfnm\relax \def\xfnm[#1]{\unskip,\space#1}\fi
%Type = Misc
\bibitem[{Cen et~al.(2025)Cen, Liu, Li \& Wang}]{cen2025sqlfixagentsemanticaccuratetexttosqlparsing}
\bibinfo{author}{Cen, J.}, \bibinfo{author}{Liu, J.}, \bibinfo{author}{Li, Z.}, \& \bibinfo{author}{Wang, J.} (\bibinfo{year}{2025}).
\newblock \bibinfo{title}{Sqlfixagent: Towards semantic-accurate text-to-sql parsing via consistency-enhanced multi-agent collaboration}.
\newblock \URLprefix \url{https://arxiv.org/abs/2406.13408}. \href{http://arxiv.org/abs/2406.13408}{\tt arXiv:2406.13408}.
%Type = Misc
\bibitem[{Chen et~al.(2023)Chen, Lin, Schärli \& Zhou}]{chen2023teachinglargelanguagemodels}
\bibinfo{author}{Chen, X.}, \bibinfo{author}{Lin, M.}, \bibinfo{author}{Schärli, N.}, \& \bibinfo{author}{Zhou, D.} (\bibinfo{year}{2023}).
\newblock \bibinfo{title}{Teaching large language models to self-debug}.
\newblock \URLprefix \url{https://arxiv.org/abs/2304.05128}. \href{http://arxiv.org/abs/2304.05128}{\tt arXiv:2304.05128}.
%Type = Misc
\bibitem[{Gao et~al.(2023)Gao, Wang, Li, Sun, Qian, Ding \& Zhang}]{gao2023texttosql}
\bibinfo{author}{Gao, D.}, \bibinfo{author}{Wang, H.}, \bibinfo{author}{Li, Y.}, \bibinfo{author}{Sun, X.}, \bibinfo{author}{Qian, Y.}, \bibinfo{author}{Ding, B.}, \& \bibinfo{author}{Zhang, J.} (\bibinfo{year}{2023}).
\newblock \bibinfo{title}{Text-to-sql empowered by large language models: A benchmark evaluation}.
\newblock \URLprefix \url{https://arxiv.org/abs/2308.15363}. \href{http://arxiv.org/abs/2308.15363}{\tt arXiv:2308.15363}.
%Type = Article
\bibitem[{Hochreiter \& Schmidhuber(1997)}]{lstm-10.1162/neco.1997.9.8.1735}
\bibinfo{author}{Hochreiter, S.}, \& \bibinfo{author}{Schmidhuber, J.} (\bibinfo{year}{1997}).
\newblock \bibinfo{title}{Long short-term memory}.
\newblock {\it \bibinfo{journal}{Neural Computation}\/},  {\it \bibinfo{volume}{9}\/}, \bibinfo{pages}{1735--1780}. \URLprefix \url{https://doi.org/10.1162/neco.1997.9.8.1735}. \DOIprefix\doi{10.1162/neco.1997.9.8.1735}. \href{http://arxiv.org/abs/https://direct.mit.edu/neco/article-pdf/9/8/1735/813796/neco.1997.9.8.1735.pdf}{\tt arXiv:https://direct.mit.edu/neco/article-pdf/9/8/1735/813796/neco.1997.9.8.1735.pdf}.
%Type = Article
\bibitem[{Li \& Jagadish(2014)}]{10.14778/2735461.2735468}
\bibinfo{author}{Li, F.}, \& \bibinfo{author}{Jagadish, H.~V.} (\bibinfo{year}{2014}).
\newblock \bibinfo{title}{Constructing an interactive natural language interface for relational databases}.
\newblock {\it \bibinfo{journal}{Proc. VLDB Endow.}\/},  {\it \bibinfo{volume}{8}\/}, \bibinfo{pages}{73–84}. \URLprefix \url{https://doi.org/10.14778/2735461.2735468}. \DOIprefix\doi{10.14778/2735461.2735468}.
%Type = Misc
\bibitem[{Liu et~al.(2023)Liu, Hu, Wen \& Yu}]{liu2023comprehensive}
\bibinfo{author}{Liu, A.}, \bibinfo{author}{Hu, X.}, \bibinfo{author}{Wen, L.}, \& \bibinfo{author}{Yu, P.~S.} (\bibinfo{year}{2023}).
\newblock \bibinfo{title}{A comprehensive evaluation of chatgpt's zero-shot text-to-sql capability}.
\newblock \URLprefix \url{https://arxiv.org/abs/2303.13547}. \href{http://arxiv.org/abs/2303.13547}{\tt arXiv:2303.13547}.
%Type = Misc
\bibitem[{Rajkumar et~al.(2022)Rajkumar, Li \& Bahdanau}]{rajkumar2022evaluatingtexttosqlcapabilitieslarge}
\bibinfo{author}{Rajkumar, N.}, \bibinfo{author}{Li, R.}, \& \bibinfo{author}{Bahdanau, D.} (\bibinfo{year}{2022}).
\newblock \bibinfo{title}{Evaluating the text-to-sql capabilities of large language models}.
\newblock \URLprefix \url{https://arxiv.org/abs/2204.00498}. \href{http://arxiv.org/abs/2204.00498}{\tt arXiv:2204.00498}.
%Type = Misc
\bibitem[{Xie et~al.(2024)Xie, Wu \& Zhou}]{xie2024magsqlmultiagentgenerativeapproach}
\bibinfo{author}{Xie, W.}, \bibinfo{author}{Wu, G.}, \& \bibinfo{author}{Zhou, B.} (\bibinfo{year}{2024}).
\newblock \bibinfo{title}{Mag-sql: Multi-agent generative approach with soft schema linking and iterative sub-sql refinement for text-to-sql}.
\newblock \URLprefix \url{https://arxiv.org/abs/2408.07930}. \href{http://arxiv.org/abs/2408.07930}{\tt arXiv:2408.07930}.

\end{thebibliography}
	
\end{document}